\documentclass{article} 

\usepackage{amssymb,amsmath,wasysym,amsthm}
\usepackage{amsmath, graphicx, enumerate,color, wasysym,amsthm,subfigure}
\usepackage{amsmath, amsthm, amssymb}

\def\supp{\operatorname{supp}}

\newcommand{\X}{\mathcal{X}}

\newcommand{\C}{\mathcal{C}}
\newcommand{\M}{\mathcal{M}}
\newcommand{\A}{\mathcal{A}}
\newcommand{\U}{\mathcal{U}}

\newcommand{\EE}{\mathbb{E}}
\newcommand{\RR}{\mathbb{R}}

\newcommand{\NN}{\mathbb{N}}

\newtheorem{theorem}{Theorem}[section]
\newtheorem{lemma}[theorem]{Lemma}

\title{Learning Probability Measures with respect to Optimal Transport Metrics}

\author{
Guille D. Canas$^{\star,\dagger}$ \hspace*{1in} Lorenzo A. Rosasco$^{\star,\dagger}$ \\
$\star$ LCSL - Massachusetts Institute of Technology\\ 
$\dagger$ Italian Institute of Technology\\ 
\texttt{\{guilledc,lrosasco\}@mit.edu}
}

\begin{document}

\maketitle

\begin{abstract}

We study the problem of estimating, in the sense of optimal transport metrics,  
	a measure which is assumed supported on a manifold embedded in a Hilbert space. 
By establishing a precise connection between optimal transport metrics, optimal quantization, and learning theory, 
	we derive new probabilistic bounds for the performance of a classic algorithm in unsupervised learning (k-means), 
	when used to produce a probability measure derived from the data. 
In the course of the analysis, we arrive at new lower bounds, as well as probabilistic upper bounds on the convergence rate of the empirical law of large numbers, 
	which, unlike existing bounds, are applicable to a wide class of measures.

\end{abstract}

\section{Introduction and Motivation}

In this paper we study the problem of learning from random samples  a probability distribution supported on a manifold,
when   the learning error is measured using transportation metrics. 


The problem of learning a probability distribution is classic in statistics and machine learning, and 
is typically analyzed for distributions in $\X=\RR^d$ that have a density with respect to the Lebesgue measure, 
with total variation, and $L_2$ among the common distances used to measure closeness of two densities
(see for instance~\cite{dev01,tsy09} and references therein.)
The setting in which the data 
distribution is supported on a low dimensional manifold embedded in a high dimensional space  has only been considered more recently. 
In particular, kernel density estimators on manifolds have been  described in \cite{vin03}, and their  pointwise consistency, as well 
as convergence rates, have been studied in \cite{pel05,ozakin09, Henry09}. 
A discussion on several topics  related to statistics on a Riemannian manifold can be found in \cite{Pen06}.

In this paper, we consider  the problem of  estimating, in the 2-Wasserstein sense, 
 a distribution  supported on a manifold embedded in a Hilbert space.
 The exact formulation  of the problem,  as well as  a detailed discussion of related previous works are given in Section~\ref{sec:setup}.

Interestingly, the problem of approximating measures with respect to transportation distances
	has deep connections with the 
fields of optimal quantization~\cite{GL00,G04}, optimal transport~\cite{V09} and, as we point out in this work, 
with 
unsupervised learning (see Sec.~\ref{sec:algos}.)
In fact, as described in the sequel, some of the most widely-used algorithms for unsupervised learning, such as k-means
	(but also others such as PCA and k-flats), can be shown
		to be performing exactly the task of estimating the data-generating measure in the sense of the 2-Wasserstein distance. 
This close relation between  learning theory, and optimal transport and quantization seems novel and of interest in its own right. 
Indeed, in this work, techniques from the above three fields are used to derive the new probabilistic bounds described below. 

Our technical contribution can be summarized as follows:
\begin{enumerate}[(a)]
\item 
	we prove uniform lower bounds for the distance between a measure and estimates based on discrete sets 
		(such as the empirical measure or measures derived from algorithms such as k-means);
\item 
	we provide new probabilistic bounds for the rate of convergence of the empirical law of large numbers
		which, unlike existing probabilistic bounds, hold for a very large class of measures;
\item 
	we provide probabilistic bounds for the rate of convergence of measures derived from k-means to the data measure. 
\end{enumerate}

The structure of the paper is described at  the end of Section~\ref{sec:setup}, where we discuss 
the exact formulation  of the problem as well as related previous works.

\section{Setup and Previous work}\label{sec:setup}

Consider the problem of learning a probability measure $\rho$ defined on a space $\M$, from an i.i.d.\ sample $X_n=(x_1,\dots,x_n)\sim\rho^n$ of size $n$. 
We assume $\M$ to be a compact, smooth d-dimensional manifold with $\C^1$ metric and volume measure $\lambda_\M$, 
	embedded in the unit ball of a separable Hilbert space $\X$ with inner product $\left< \cdot,\cdot \right>$, 
	induced norm $\|\cdot\|$, and distance $d$
(for instance $\M=B_2^d(1)$ the unit ball in $\X=\RR^d$.)
Following~\cite[p.\ 94]{V09}, let $P_p(\M)$ denote the Wasserstein space of order $1\le p < \infty$:
\[ P_p(\M) := \left\{   \rho \in P(\M) : \displaystyle{\int_\M \|x\|^p d\rho(x) < \infty }  \right\} \]
of probability measures with finite p-th moment. 
The p-Wasserstein distance
\begin{equation}\label{eq:Wass}
 W_p(\rho,\mu) = \inf\left\{ \left[ \EE \|X-Y\|^p \right]^{1/p} , \text{ Law}(X) = \rho, \text{ Law}(Y) = \mu \right\} 
\end{equation}
where the $inf$ is over random variables $X,Y$ with laws $\rho,\mu$, respectively, 
is the optimal expected cost of transporting points generated from $\rho$ to those generated from $\mu$, 
	and is guaranteed to be finite in $P_p(\M)$ \cite[p.\ 95]{V09}. 
The space $P_p(\M)$ with the $W_p$ metric is itself a complete separable metric space~\cite{V09}. 
We consider here the problem of learning probability measures $\rho\in P_2(\M)$, where the performance is measured by the distance $W_2(\rho,\cdot)$. 

There are many possible choices of distances between probability measures~\cite{GS02}. 
Among them, $W_p$ metrizes weak convergence (see~\cite{V09} theorem~6.9), that is, 
	in $P_p(\M)$, a sequence $(\mu_i)_{i\in\NN}$ of measures converges weakly to $\mu$ iff $W_p(\mu_i,\mu)\rightarrow 0$.
There are other distances, 
	such as the L\'evy-Prokhorov, or the \mbox{weak-*} distance, 
	that also metrize weak convergence. 
However, as pointed out by Villani in his excellent monograph~\cite[p.\ 98]{V09},
\begin{enumerate}
	\item ``Wasserstein distances are rather strong, [...]a definite advantage over the \mbox{weak-*} distance". 
	\item ``It is not so difficult to combine information on convergence in Wasserstein distance with some smoothness bound, in order to get convergence in stronger distances."
\end{enumerate}
Wasserstein distances have been used to study the mixing and convergence of Markov chains~\cite{O09}, as
	well as concentration of measure phenomena~\cite{L01}. 
To this list we would add the important fact that existing and widely-used algorithms for unsupervised learning 
	can be easily extended (see Sec.~\ref{sec:algos}) to compute a measure $\rho'$ that minimizes the distance $W_2(\hat\rho_n, \rho')$ to the empirical measure 
		\[ \hat\rho_n := \frac 1 n \displaystyle{ \sum_{i=1}^n \delta_{x_i}}, \]
a fact that will allow us to prove, in Sec.~\ref{sec:ubounds}, bounds on the convergence of the measure induced by k-means to the population measure $\rho$. 

The most useful versions of Wasserstein distance are $p=1,2$, with $p=1$ being the weaker of the two 
	(by H\"older's inequality, $p \le q \Rightarrow W_p \le W_q$; a discussion of $p=\infty$ would take us out of topic, since its behavior is markedly different.) 
In particular, ``results in $W_2$ distance are usually stronger, and more difficult to establish than results in $W_1$ distance"~\cite[p.\ 95]{V09}.

\subsection{Closeness of Empirical and Population Measures}\label{sec:rhorhon}

By the \emph{empirical law of large numbers}, the empirical measure converges almost surely to the population measure: $\hat\rho_n\rightarrow\rho$ 
	in the sense of the weak topology~\cite{V58}.
Since weak convergence and convergence in $W_p$ are equivalent in $P_p(\M)$, 
	this means that, in the $W_p$ sense, the empirical measure $\hat\rho_n$ is an arbitrarily good approximation of $\rho$, as $n\rightarrow\infty$. 
A fundamental question is therefore how fast the rate of convergence of $\hat\rho_n\rightarrow\rho$ is.

\subsubsection{Convergence in expectation}
The \emph{mean} rate of convergence of $\hat\rho_n\rightarrow\rho$ has been widely studied in the past, 
	resulting in upper bounds of order $\EE W_2(\rho,\hat\rho_n) = O(n^{-1/(d+2)})$~\cite{HK94,CCDM11}, 
	and lower bounds of order $\EE W_2(\rho,\hat\rho_n) = \Omega(n^{-1/d})$~\cite{R91} 
	(both assuming that the absolutely continuous part of $\rho$ is $\rho_A\ne 0$, with possibly better rates otherwise). 
	
More recently, an upper bound of order $\EE W_p(\rho,\hat\rho_n) = O(n^{-1/d})$ has been proposed~\cite{BB11} by 
	proving a bound for the Optimal Bipartite Matching (OBM) problem~\cite{AKT84}, and relating this problem to the expected distance $\EE W_p(\rho,\hat\rho_n)$. 
In particular, given two independent samples $X_n, Y_n$, 
	the OBM problem is that of finding a permutation $\sigma$ 
	that minimizes the matching cost $n^{-1}\sum \| x_i - y_{\sigma(i)}\|^p$~\cite{P78,S97}. 
It is not hard to show that the optimal matching cost is $W_p(\hat\rho_{_{X_n}}, \hat\rho_{_{Y_n}})^p$,  
	where $\hat\rho_{_{X_n}},\hat\rho_{_{Y_n}}$ are the empirical measures associated to $X_n,Y_n$. 
By Jensen's inequality, the triangle inequality, and $(a+b)^p \le 2^{p-1} (a^p + b^p)$, it holds
	\[ \EE W_p(\rho,\hat\rho_n)^p \le \EE W_p(\hat\rho_{_{X_n}},\hat\rho_{_{Y_n}})^p \le 2^{p-1} \EE W_p(\rho,\hat\rho_n)^p, \]
and therefore a bound of order $O(n^{-p/d})$ for the OBM problem~\cite{BB11} implies a bound $\EE W_p(\rho,\hat\rho_n) = O(n^{-1/d})$. 
The matching lower bound is only known for a special case: $\rho_A$ constant over a bounded set of non-null measure~\cite{BB11} (e.g.\ $\rho_A$ uniform.)
Similar results, with matching lower bounds are found for $W_1$ in~\cite{DY95}. 

\subsubsection{Convergence in probability}

Results for convergence in probability, one of the main results of this work, appear to be considerably harder to obtain. 
One fruitful avenue of analysis has been the use of so-called \emph{transportation}, or \emph{Talagrand inequalities} $T_p$, which can be used to prove concentration inequalities on $W_p$ \cite{L01}. 
In particular, we say that $\rho$ satisfies a $T_p(C)$ inequality with $C>0$
	iff $W_p(\rho,\mu)^2 \le C H(\mu|\rho), \forall\mu\in P_p(\M)$, where $H(\cdot | \cdot)$ is the relative entropy~\cite{L01}.  
As shown in~\cite{BGV07,B11}, it is possible to obtain probabilistic upper bounds on $W_p(\rho,\hat\rho_n)$, with $p=1,2$, if $\rho$ is known to satisfy a $T_p$ inequality
	of the same order, thereby reducing the problem of bounding $W_p(\rho,\hat\rho_n)$ to that of obtaining a $T_p$ inequality. 
Note that, by Jensen's inequality, and as expected from the behavior of $W_p$, the inequality $T_2$ is stronger than $T_1$~\cite{L01}. 

While it has been shown that $\rho$ satisfies a $T_1$ inequality iff it has a finite square-exponential moment~\cite{BG99,BV05}, 
	no such general conditions have been found for $T_2$. 
As an example, consider that, if $\M$ is compact with diameter $D$ then, by theorem~6.15 of~\cite{V09}, 
	and the celebrated Csisz\'ar-Kullback-Pinsker inequality~\cite{P64}, for all $\rho,\mu\in P_p(\M)$, it is
	\[ W_p(\rho,\mu)^{2p} \le (2D)^{2p} \| \rho - \mu \|^2_{\text{TV}} \le 2^{2p-1} D^{2p} H(\mu | \rho), \] 
where $\|\cdot\|_{\text{TV}}$ is the total variation norm. 
Clearly, this implies a $T_{p=1}$ inequality, but for $p\ge 2$ it does not. 

The $T_2$ inequality has been shown by Talagrand 
	to be satisfied by the Gaussian distribution~\cite{T96}, and then slightly more generally by strictly log-concave measures~\cite{B03}. 
However, as noted in~\cite{BGV07}, ``contrary to the $T_1$ case, there is no hope to obtain $T_2$ inequalities from just integrability or decay estimates."

\noindent{\bf Structure of this paper}. 
In this work we obtain bounds in probability (learning rates) for the problem of learning a probability measure (in the sense of $W_2$.)
We begin by establishing (lower) bounds for the convergence of empirical to population measures, 
	which serve to set up the problem and introduce the connection between quantization and measure learning (sec.~\ref{sec:LPM}.)
We then describe how existing unsupervised learning algorithms that compute a set (k-means, k-flats, PCA,\dots) can be easily extended to produce a measure
(sec.~\ref{sec:algos}.) Due to its simplicity and widespread use, we focus here on k-means. 
Since the two measure estimates that we consider are the empirical measure, and the measure induced by k-means, we next set out to prove upper bounds on 
	their convergence to the data-generating measure (sec.~\ref{sec:ubounds}.)
We arrive at these bounds by means of intermediate measures, which are related to the problem of optimal quantization. 
The bounds apply in a very broad setting (unlike existing bounds based on transportation inequalities, they are not restricted to log-concave measures.)

\section{Learning probability measures, optimal transport and quantization}\label{sec:LPM}

We address the problem of learning a probability measure $\rho$ when the only observation we have at our disposal is an i.i.d.\ sample $X_n$. 
We begin by establishing some notation and useful intermediate results. 

Given a closed set $S\subseteq\M$, let $\pi_{_S} = \sum_{q\in S} \mathbf{1}_{V_q(S)} \cdot q$ be a  
	nearest neighbor projection onto $S$ (a function mapping points in $\X$ to their closest point in $S$), where $\{V_q(S) : q\in S\}$ is a Borel Voronoi partition of $\X$
	such that $V_q(S) \subseteq \{ x\in\X : \|x-q\| = \min_{r\in S}\|x - r\| \}$ (see for instance~\cite{GLP08}.) 
Since $S$ is closed and $\|x-\cdot\|$ is continuous and convex, every points $x\in\X$ has a closest point in $S$. 
	Since $\{V_q(S) : q\in S\}$ is a Borel partition, it follows that $\pi_{_S}$ is a measurable map. 
For any $\rho\in P_p(\M)$, the pushforward, or image measure $\pi_{_S}\rho$ under the mapping $\pi_{_S}$ 
	is supported in $S$, and is such that, for Borel measurable sets $A$, it is $(\pi_{_S}\rho)(A) := \rho(\pi^{-1}_S(A))$. 

%


We now establish a connection between the expected distance to a set $S$, and the distance between $\rho$ and the set's induced pushforward measure.
Notice that the expected distance to $S$ is exactly the expected quantization error incurred when encoding points drawn from $\rho$ by their closest point in $S$. 
This close connection between optimal quantization and Wasserstein distance has been pointed out in the past in the 
	statistics~\cite{P82}, optimal quantization~\cite[p.\ 33]{GL00}, and approximation theory literatures~\cite{G04}. 

The following two lemmas are key tools in the reminder of the paper. The first highlights the close link between quantization and optimal transport. 

\begin{lemma}\label{lem:EEW}
	For closed $S\subseteq\M$, $\rho\in P_p(\M)$, $1\le p<\infty$, it holds
		$ \EE_{x\sim\rho} d(x, S)^p = W_p(\rho, \pi_{_S}\rho)^p$.
\end{lemma}

Note that the key element in the above lemma is that the two measures in the expression $W_p(\rho,\pi_{_S}\rho)$ must match. 
When there is a mismatch, the distance can only increase. 
That is, $W_p(\rho, \pi_{_S}\mu) \ge W_p(\rho, \pi_{_S}\rho)$ for all $\mu\in P_p(\M)$. 
In fact, the following lemma shows that, 
	among all the measures with support in $S$, 
	$\pi_{_S}\rho$ is closest to $\rho$. 

\begin{lemma}\label{lem:WW2}
	For closed $S$, and all $\mu\in P_p(\M)$ with  $\supp(\mu)\subseteq S$, $1\le p<\infty$, it holds
		$ W_p(\rho, \mu) \ge W_p(\rho, \pi_{_S}\rho) $.
\end{lemma}

When combined, lemmas~\ref{lem:EEW} and~\ref{lem:WW2} indicate that the behavior of the measure learning problem 
	is limited by the performance of the optimal quantization problem. 
For instance, $W_p(\rho,\hat\rho_n)$ can only be, in the best-case, as low as the optimal quantization cost with codebook of size $n$. 
The following section makes this claim precise. 




\subsection{Lower bounds}\label{sec:lbounds}

Consider the situation depicted in fig.~\ref{fig:pizza}, in which a sample $X_4=\{x_1,x_2,x_3,x_4\}$ is drawn from a distribution $\rho$ 
	which we assume here to be absolutely continuous on its support.  
As shown, the projection map $\pi_{_{X_4}}$ sends points $x$ to their closest point in $X_4$. 
The resulting Voronoi decomposition of $\supp(\rho)$ is drawn in shades of blue.  
By lemma~5.2 of~\cite{C06}, the pairwise intersections of Voronoi regions have null ambient measure, 
	and since $\rho$ is absolutely continuous, 
	the pushforward measure can be written in this case as $\pi_{_{X_4}}\rho = \sum_{j=1}^4 \rho(V_j)\delta_{x_j} $, 
	where $V_j$ is the Voronoi region of $x_j$. 
Note that this decomposition is not always possible if, for instance $\rho$ has an atom falling on two Voronoi regions: 
	both regions would count the atom as theirs, and double-counting would imply $\sum_j \rho(V_j) > 1$. 
The technicalities required to correctly define $\rho(V_j)$ are such that, 
	in general,  it is simpler to write $\pi_{S}\rho$, even though (if $S$ is discrete) this measure can clearly be written as a sum of deltas with appropriate masses. 

By lemma~\ref{lem:EEW}, the distance $W_p(\rho,\pi_{_{X_4}}\rho)^p$ is the 
	(expected) quantization cost of $\rho$ when using $X_4$ as codebook. 
Clearly, this cost can never be lower than the \emph{optimal} quantization cost of size $4$. This reasoning leads to the following lower bound between empirical and population measures.

\begin{theorem}
\label{th:lbound}
	For $\rho\in P_p(\M)$ with absolutely continuous part $\rho_A\ne 0$, and $1\le p < \infty$, it holds
		$ W_p(\rho, \hat\rho_n) = \Omega(n^{-1/d}) $ 
	uniformly over $\hat\rho_n$, where the constants depend on $d$ and $\rho_A$ only. 
\end{theorem}
\noindent\emph{Proof:} 	
Let $V_{n,p}(\rho) := \inf_{S\subset\M, |S|= n} \EE_{x\sim\rho} d(x, S)^p$ be the optimal quantization cost of $\rho$ of order $p$ with $n$ centers. 
Since $\rho_A\ne 0$, and since $\rho$ has a finite $(p+\delta)$-th order moment, for some $\delta >0$ (since it is supported on the unit ball), then it is $V_{n,p}(\rho) = \Theta(n^{-p/d})$, with constants depending on $d$ and $\rho_A$ (see~\cite[p.\ 78]{GL00} and~\cite{G04}.) 
Since $\supp(\hat\rho_n) = X_n$, it follows that
	\[ W_p( \rho, \hat\rho_n)^p \underset{\text{lemma~\ref{lem:WW2}}}{\ge}  W_p(\rho, \pi_{_{X_n}}\rho)^p 
		\underset{\text{lemma~\ref{lem:EEW}}}{=} \EE_{x\sim\rho} d(x, X_n)^p \ge V_{n,p}(\rho) =  \Theta(n^{-p/d}) \qed \] 
		
Note that the bound of theorem~\ref{th:lbound} holds for $\hat\rho_n$ derived from \emph{any} sample $X_n$, 
	and is therefore stronger than the existing lower bounds on the 
		convergence rates of $\EE W_p(\rho,\hat\rho_n)\rightarrow 0$. 
In particular, it trivially induces the known lower bound $\Omega(n^{-1/d})$ on the expected rate of convergence. 

The consequence of theorem~\ref{th:lbound} is clearly that the rate of convergence of the empirical law of large numbers 
	is limited (in all cases), by the dimension of the space in which $\rho$ is absolutely continuous. 
This justifies the choice of formal setting to be a $d$-manifold (or even $\mathbb{R}^d$): 
	by the above uniform lower bound, 
		one is effectively forced to make a finite-dimension assumption on the space where $\rho$ is absolutely continuous. 

\section{{Unsupervised learning algorithms for  learning a probability measure}}\label{sec:algos}

As described in~\cite{MP10}, 
	several of the most widely used unsupervised learning algorithms can be interpreted to take as input a sample $X_n$ and output a set $\hat{S}_k$, 
		where $k$ is typically a free parameter of the algorithm, 
		such as the number of means in k-means\footnote{
			In a slight abuse of notation, we refer to the k-means algorithm here as an ideal algorithm that solves the k-means problem, 
				even though in practice an approximation algorithm may be used.
		}, the dimension of affine spaces in PCA, etc.
Performance is measured by the empirical quantity 
	$n^{-1}\sum_{i=1}^n d(x_i , \hat{S}_k)^2$, which is minimized among all sets in some class (e.g.\ sets of size $k$, affine spaces of dimension $k$,\dots)
This formulation is general enough to encompass k-means and PCA, but also k-flats, 
	non-negative matrix factorization, 
	and  sparse coding 
(see~\cite{MP10} and references therein.)

Using the discussion of Sec.~\ref{sec:LPM}, we can establish a clear connection between unsupervised learning
	and the problem of learning probability measures with respect to $W_2$. 
Consider as a running example the k-means problem, though the argument is general. 
Given an input $X_n$, the k-means problem is to find a set $|\hat{S}_k|=k$ 
	minimizing its average distance from points in $X_n$. 
By associating to $\hat{S}_k$ the pushforward measure $\pi_{\hat{S}_k}\hat{\rho}_n$, we find that 
	\begin{equation}\label{eq:empKM}
		 \frac 1 n \sum_{i=1}^n d(x_i , \hat{S}_k)^2 
		 	= \EE_{x\sim\hat\rho_n} d(x,\hat{S}_k)^2 
			\underset{\text{lemma~\ref{lem:EEW}}}{=} W_2(\hat\rho_n, \pi_{\hat{S}_k}\hat{\rho}_n)^2. 
	\end{equation}
Since k-means minimizes equation~\ref{eq:empKM}, it also finds  
the measure that is closest to $\hat\rho_n$, among those with support of size $k$. 
This connection between k-means and $W_2$ measure approximation was, to the best of the authors' knowledge, 
	first suggested by Pollard~\cite{P82} though, as mentioned earlier, the argument carries over to many other unsupervised learning algorithms. 
	
We briefly clarify the steps involved in using an existing unsupervised learning algorithm for probability measure learning. 
Let $\U_k$ be a parametrized algorithm (e.g.\ k-means) that takes a sample $X_n$ and outputs a set $\U_k(X_n)$. The measure learning algorithm 
 $\A_k:\M^n\rightarrow P_p(\M)$ corresponding to $\U_k$ is defined as follows:
	\begin{enumerate}
		\item  $\A_k$ takes a sample $X_n$ and outputs the measure $\pi_{\hat{S}_k}\hat\rho_n$, supported on $\hat{S}_k = \U_k(X_n)$; 
		\item since $\hat\rho_n$ is discrete, then so must $\pi_{\hat{S}_k}\hat\rho_n$ be, and thus  $\A_k(X_n) = \frac 1 n \sum_{i=1}^n \delta_{\pi_{\hat{S}_k}(x_i)}$;  
		\item in practice, we can simply store an $n$-vector $\left[ \pi_{\hat{S}_k}(x_1), \dots, \pi_{\hat{S}_k}(x_n) \right]$, 
				from which $\A_k(X_n)$ can be reconstructed by placing atoms of mass $1/n$ at each point. 
	\end{enumerate}
In the case that $\U_k$ is the k-means algorithm, only $k$ points 	and $k$ masses need to be stored. 
	
Note that any algorithm $\A'$ that attempts to output a measure $\A'(X_n)$ close to $\hat\rho_n$ can be cast in the above framework. 
Indeed, if $S'$ is the support of $\A'(X_n)$ then, by lemma~\ref{lem:WW2}, $\pi_{S'}\hat\rho_n$ is the measure closest to $\hat\rho_n$ 
with support in $S'$. This effectively reduces the problem of learning a measure to that of finding a set, and is akin to how 
	the fact that every optimal quantizer is a nearest-neighbor quantizer (see~\cite{GLP08}, \cite[p.\ 350]{GG92}, and~\cite[p.\ 37--38]{GL00}) reduces the problem of finding an optimal quantizer to that of finding an optimal quantizing \emph{set}. 

Clearly, the minimum of equation~\ref{eq:empKM} 
	over sets of size $k$ (the output of k-means) is monotonically non-increasing with $k$. 
	In particular, since $\hat{S}_n = X_n$ and $\pi_{\hat{S}_n}\hat{\rho}_n = \hat\rho_n$, it is 
	$\EE_{x\sim\hat\rho_n} d(x,\hat{S}_n)^2 = W_2(\hat\rho_n, \pi_{\hat{S}_n}\hat{\rho}_n)^2=0$. 
	That is, we can always make the learned measure arbitrarily close to $\hat\rho_n$ by increasing $k$. 
However, as pointed out in Sec.~\ref{sec:setup}, the problem of measure learning is concerned with minimizing the distance $W_2(\rho,\cdot)$ 
	to the data-generating measure. 
The actual performance of k-means is thus not necessarily guaranteed to behave in the same way as the empirical one, 
	and the question of characterizing its behavior 
	as a function of $k$ and $n$ naturally arises. 
	
Finally, we note that, while it is $\EE_{x\sim\hat\rho_n} d(x,\hat{S}_k)^2 = W_2(\hat\rho_n, \pi_{\hat{S}_k}\hat{\rho}_n)^2$ 
	(the empirical performances are the same in the optimal quantization, 
	and measure learning problem formulations), 
	the actual performances satisfy
	\[ \EE_{x\sim\rho} d(x,\hat{S}_k)^2 \underset{\text{lemma~\ref{lem:EEW}}}{=} 
		W_2(\rho,\pi_{\hat{S}_k}\rho)^2 \underset{\text{lemma~\ref{lem:WW2}}}{\le} W_2(\rho, \pi_{\hat{S}_k}\hat{\rho}_n)^2, \quad 1\le k\le n. \]
Consequently, with the identification between sets $S$ and measures $\pi_{_S}\hat\rho_n$, 
	the set-approximation problem is, in general, \emph{different} from the measure learning problem 
		(for example, if $\M=\RR^d$ and $\rho$ is absolutely continuous over a set of non-null volume, 
			 it's not hard to show that the inequality is almost surely strict: $ \EE_{x\sim\rho} d(x,\hat{S}_k)^2 
			 	< W_2(\rho, \pi_{\hat{S}_k}\hat{\rho}_n)^2$ for $n>k>1$.)

In the remainder, we characterize the performance of k-means on the measure learning problem, for varying $k,n$. 
Although other unsupervised learning algorithms could have been chosen as basis for our analysis, k-means is one of the oldest and most widely used, 
	and the one for which the deep connection between optimal quantization and measure approximation is most clearly manifested. 
Note that, by setting $k=n$, our analysis includes the problem of characterizing the behavior of the distance $W_2(\rho,\hat\rho_n)$ 
	between empirical and population measures which, as indicated in Sec.~\ref{sec:rhorhon}, is a fundamental question in statistics 
	(i.e.\ the speed of convergence of the empirical law of large numbers.)

\section{Learning rates}\label{sec:ubounds}

\begin{figure}[ht]
\begin{minipage}[b]{0.5\linewidth}
\centering
\includegraphics[width=6.5cm]{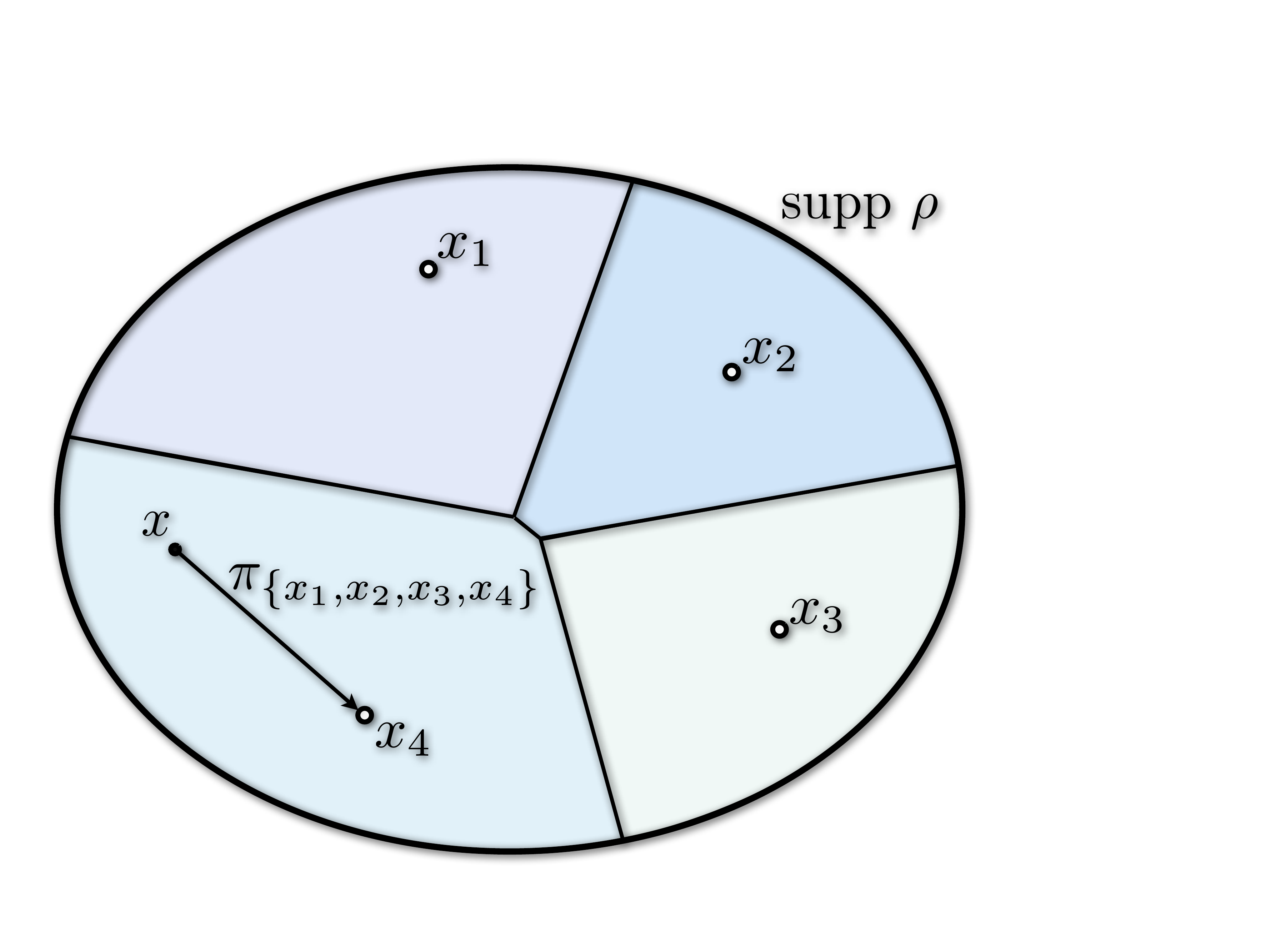}
\caption{A sample $\{x_1,x_2,x_3,x_4\}$ is drawn from a distribution $\rho$ with support in $\supp\rho$. 
The projection map $\pi_{\{x_1,x_2,x_3,x_4\}}$ sends points $x$ to their closest one in the sample. 
The induced Voronoi tiling is shown in shades of blue. 
}
\label{fig:pizza}
\end{minipage}
\hspace{0.5cm}
\begin{minipage}[b]{0.5\linewidth}
\centering
\includegraphics[width=6.5cm]{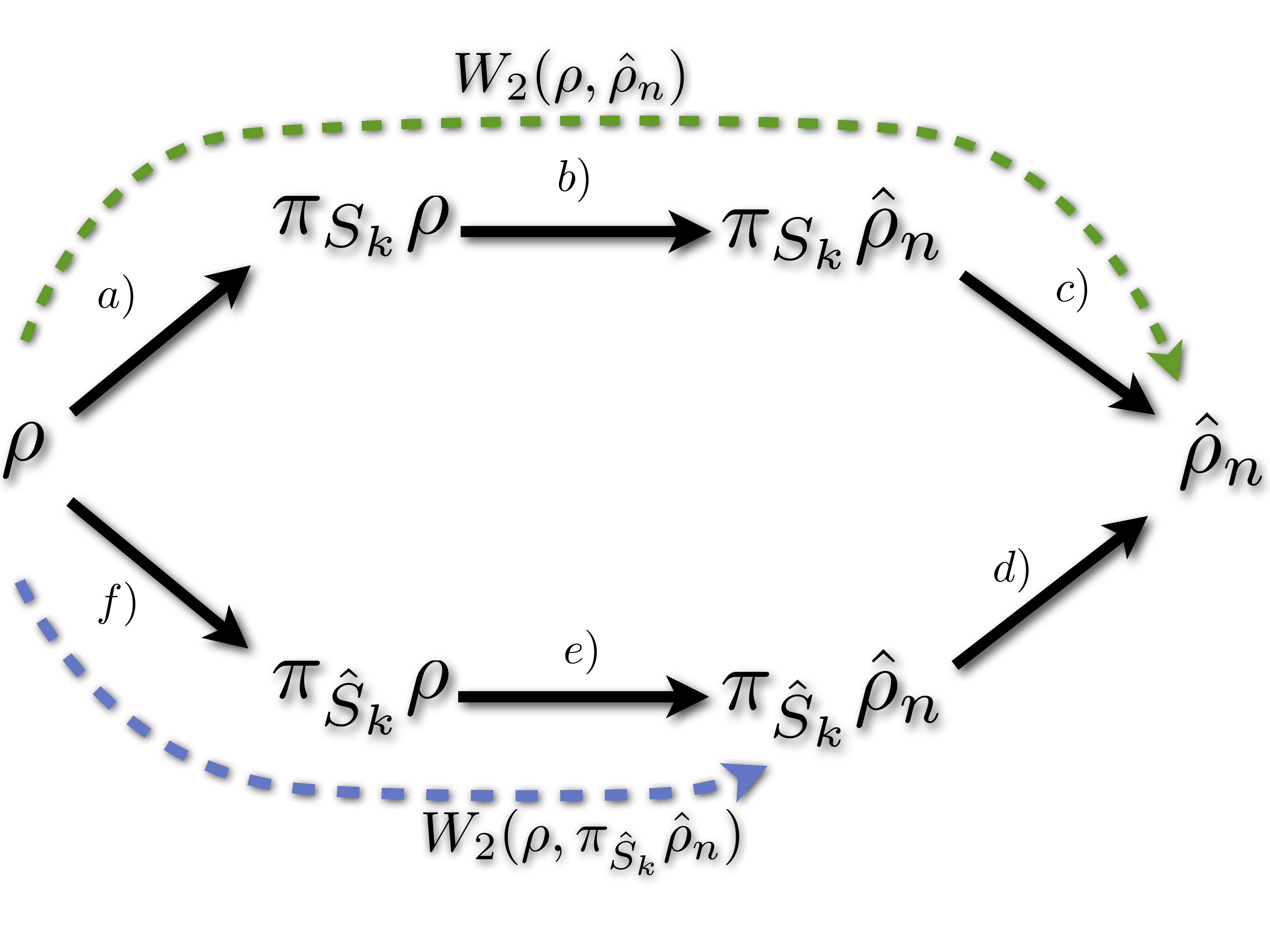}
\caption{The measures considered in this paper are linked by arrows for which upper bounds
for their distance are derived. 
Bounds for the quantities of interest $W_2(\rho,\hat\rho_n)^2$, and $W_2(\rho,\pi_{\hat{S}_k}\hat\rho_n)^2$, are decomposed by following the top and bottom colored arrows. 
}
\label{fig:benzene}
\end{minipage}
\end{figure}

In order to analyze the performance of k-means as a measure learning algorithm, 
	and the convergence of empirical to population measures, 
	we propose the decomposition shown in fig.~\ref{fig:benzene}. %
The diagram includes all the measures considered in the paper, and shows the two decompositions used to prove upper bounds. 
The upper arrow (green), illustrates the decomposition used to bound the distance $W_2(\rho,\hat\rho_n)$, 
This decomposition uses 
	the  measures $\pi_{S_k}\rho$ and $\pi_{S_k}\hat\rho_n$ as intermediates to arrive at $\hat\rho_n$, where $S_k$ is a $k$-point optimal quantizer of $\rho$, that is, a set $S_k$ minimizing $\EE_{x\sim\rho} d(x,S)^2$ and such that $|S_k|=k$.
The lower arrow (blue) corresponds to the decomposition of $W_2(\rho, \pi_{\hat{S}_k}\hat\rho_n)$ (the performance of k-means), 
whereas the labelled black arrows correspond to individual terms in the bounds. 
We begin with the (slightly) simpler of the two results.

\subsection{Convergence rates for the empirical law of large numbers }

Let $S_k$ be the optimal $k$-point quantizer of $\rho$ of order two~\cite[p.\ 31]{GL00}.
By the triangle inequality and the identity $(a+b+c)^2\le 3(a^2+b^2+c^2)$, it follows that
	\begin{equation}\label{eq:UPdecomp}
		 W_2(\rho,\hat\rho_n)^2 \le 3 \left[  W_2(\rho,\pi_{S_k}\rho)^2 + W_2(\pi_{S_k}\rho,\pi_{S_k}\hat\rho_n)^2 + W_2(\pi_{S_k}\hat\rho_n, \hat\rho_n)^2 \right].
	\end{equation}
This is the decomposition depicted in the upper arrow of fig.~\ref{fig:benzene}. 

By lemma~\ref{lem:EEW}, the first term in the sum of equation~\ref{eq:UPdecomp} is the optimal $k$-point quantization error of $\rho$ 
 over a $d$-manifold $\M$ which, using recent techniques from~\cite{G04} (see also~\cite[p.\ 491]{G07}), is shown in the proof of theorem~\ref{th:UPemp} (part a) to be of order $\Theta(k^{-2/d})$. 
The remaining terms, $b)$ and $c)$, are slightly more technical and are bounded in the proof of theorem~\ref{th:UPemp}. 

Since equation~\ref{eq:UPdecomp} holds for all $1 \le k \le n$, the best bound on $W_2(\rho, \hat\rho_n)$ 
	can be obtained by optimizing the right-hand side 
	over all possible values of $k$, 
	resulting in the following probabilistic bound for the rate of convergence of the empirical law of large numbers. 
		
\begin{theorem}\label{th:UPemp}
	Given $\rho\in P_p(\M)$ with absolutely continuous part $\rho_A\ne 0$,  sufficiently large $n$, 
	and $0 < \delta < 1$, 
		it holds 
		\[  
			W_2(\rho, \hat\rho_n) \le C \cdot m(\rho_A)  \cdot n^{-1/(2d+4)}  \cdot\tau, 
			\quad\text{ with probability }1-e^{-\tau^2}. 
		\]
	where $m(\rho_A):=\int_{\M} \rho_A(x)^{d/(d+2)} d\lambda_\M(x)$, and $C$ depends only on $d$. 
\end{theorem}
\begin{proof}See Appendix.\end{proof}

\subsection{Learning rates of k-means}

The key element in the proof of theorem~\ref{th:UPemp} is that the distance between population and empirical measures 
	can be bounded by choosing an intermediate optimal quantizing measure of an appropriate size $k$. 
In the analysis, the best bounds are obtained for $k$ smaller than $n$. If the output of k-means is close to an optimal quantizer 
(for instance if sufficient data is available), then we would similarly expect that the best bounds for k-means correspond to a choice of $k < n$. 

The decomposition of the bottom (blue) arrow in figure~\ref{fig:benzene} leads to the following bound in probability. 

\begin{theorem}\label{th:UPkmeans}
	Given $\rho\in P_p(\M)$ with absolutely continuous part $\rho_A\ne 0$, and $0<\delta < 1$, then for all sufficiently large $n$, 
	and letting
		\[ k = C\cdot m(\rho_A)\cdot n^{d/(2d+4)}, \]
		it holds 
		\[  
			 W_2(\rho, \pi_{\hat{S}_k}\hat\rho_n) \le C \cdot m(\rho_A)  \cdot n^{-1/(2d+4)}  \cdot\tau, 
			\quad\text{ with probability }1-e^{-\tau^2}. 
		\]
	where $m(\rho_A):=\int_{\M} \rho_A(x)^{d/(d+2)} d\lambda_\M(x)$, and $C$ depends only on $d$. 
\end{theorem}
\begin{proof}See Appendix.\end{proof}

Note that the upper bounds in theorem~\ref{th:UPemp} and~\ref{th:UPkmeans} are exactly the same. 
Although this may appear surprising, it stems from the following fact. 
Since $S=\hat{S}_k$ is a minimizer of $W_2(\pi_S\hat\rho_n, \hat\rho_n)^2$, 
	the bound d) of figure~\ref{fig:benzene} satisfies: 
\[ 
	W_2(\pi_{\hat{S}_k}\hat\rho_n, \hat\rho_n)^2 \le  W_2(\pi_{S_k}\hat\rho_n, \hat\rho_n)^2
\]
and therefore (by the definition of c), the term d) is of the same order as c). 
Since f) is also of the same order as c) (see the proof of theorem~\ref{th:UPkmeans}), 
	this means that, up to a small constant factor, 
	adding the term d) to the bound of $W_2(\rho, \pi_{\hat{S}_k}\hat\rho_n)^2$ does not affect the bound. 
Since d) is the term that takes the output measure of k-means to the empirical measure, this implies that 
	the rate of convergence of k-means cannot be worse than that of $\hat\rho_n\rightarrow\rho$. 
Conversely, bounds for  $\hat\rho_n\rightarrow\rho$ are obtained from best rates of convergence of optimal quantizers, 
	whose convergence to $\rho$ cannot be slower than that of k-means 
	(since the quantizers that k-means produces are suboptimal.)
	
Since the bounds obtained for the convergence of $\hat\rho_n\rightarrow\rho$ 
	are the same as those for k-means with $k$ of order $k=\Theta(n^{d/(2d+4)})$, 
	this implies that estimates of $\rho$ that are as accurate as those derived from an $n$ point-mass measure $\hat\rho_n$
	can be derived from $k$ point-mass measures with $k\ll n$. 

Finally, we note that the introduced bounds are currently limited by the statistical bound 
\[	
	\sup_{|S|=k} | W_2(\pi_{S}\hat\rho_n, \hat\rho_n)^2 - W_2(\pi_{S}\rho, \rho)^2| \underset{\text{lemma~\ref{lem:EEW}}}{= }
	\sup_{|S|=k} | \EE_{x\sim\hat\rho_n}d(x,S)^2 - \EE_{x\sim\rho}d(x,S)^2| 
\]
(see for instance~\cite{MP10}), for which non-matching lower bounds are known. 
This means that, if better upper bounds can be obtained, then both bounds in theorems~\ref{th:UPemp} and~\ref{th:UPkmeans}
	would automatically improve.



\bibliographystyle{plain}
{\small \bibliography{wasserstein03_GC}}
\end{document}